\documentclass[runningheads]{llncs}
\usepackage[T1]{fontenc}
\usepackage{booktabs}
\usepackage{multirow}
\usepackage{amsmath}
\usepackage{amssymb}
\usepackage{tabularx}
\usepackage{graphicx,verbatim}
\usepackage{comment}

\usepackage{subcaption}

\begin{document}
\title{MAC-XA: Multi-view Anatomy-Correspondence Fusion for Coronary Stenosis Reporting from X-ray Angiography}

% Simulation-Based Multi-view Anatomy-aligned Fusion Augments Coronary Stenosis Reporting 

% Using Digitally Reconstructed Radiographs
%
\titlerunning{MAC-XA}

\begin{comment}  %% Removed for anonymized MICCAI 2025 submission
\author{First Author\inst{1}\orcidID{0000-1111-2222-3333} \and
Second Author\inst{2,3}\orcidID{1111-2222-3333-4444} \and
Third Author\inst{3}\orcidID{2222--3333-4444-5555}}
%
\authorrunning{F. Author et al.}
\institute{Princeton University, Princeton NJ 08544, USA \and
Springer Heidelberg, Tiergartenstr. 17, 69121 Heidelberg, Germany
\email{lncs@springer.com}\\
\url{http://www.springer.com/gp/computer-science/lncs} \and
ABC Institute, Rupert-Karls-University Heidelberg, Heidelberg, Germany\\
\email{\{abc,lncs\}@uni-heidelberg.de}}

\end{comment}

\author{
Chen Jia\inst{1,6} \and
Baochang Zhang\inst{1,2,3}\thanks{Corresponding author.} \and
Fatia Kusuma Dewi\inst{1} \and
Amir Yousefi\inst{1} \and
Heribert Schunkert\inst{2,4} \and
Reza Ghotbi\inst{5} \and
Nassir Navab\inst{1,3}
}

\authorrunning{C. Jia et al.}

\institute{
Computer Aided Medical Procedures, Technical University of Munich, Munich, Germany\\
\email{\{chen.jia,baochang.zhang\}@tum.de}
\and
German Heart Center Munich, Munich, Germany
\and
Munich Center for Machine Learning, Munich, Germany
\and
German Centre for Cardiovascular Research, Munich Heart Alliance, Munich, Germany
\and
HELIOS Hospital West of Munich, Munich, Germany
\and
relAI -- Konrad Zuse School of Excellence in Reliable AI, Munich, Germany
}
\maketitle              % typeset the header of the contribution
\begin{abstract}
Multi-view reasoning in coronary X-ray angiography is inherently a cross-projection geometric problem, yet automated report generation in this setting remains largely unexplored. The 3D vascular topology leads to projection-dependent branch overlap and foreshortening, rendering single-view modeling fundamentally incomplete and unstable for lesion localization and stenosis grading. Although multi-view fusion appears promising, learning anatomically consistent fusion from real angiograms is impeded by a critical limitation: cross-view alignment is unobservable and cannot be explicitly supervised. Consequently, conventional fusion relies on implicit correlations rather than verified anatomical correspondence. We address this by reformulating multi-view stenosis reporting as an alignment-constrained aggregation problem. A controllable synthetic angiography generation strategy is introduced to expose geometry-derived patch-level correspondence supervision unavailable in real data. An anatomy-correspondence module learns cross-view correspondence matrices that explicitly align auxiliary features within the main-view coordinate space prior to fusion, thereby constraining evidence aggregation to anatomically consistent regions. Experiments on synthetic data and zero-shot transfer to real angiograms show that this alignment-constrained design improves correspondence consistency and structured stenosis reporting compared to single-view modeling and conventional multi-view fusion methods. The code will be publicly available upon publication.

% Most existing medical image report generation methods are built on single-view data, whereas multi-view acquisition is routine in real-world coronary angiography. Due to the strong 3D nature of coronary anatomy, different projections often suffer from branch overlap and occlusion. A single view therefore provides only partial evidence and leads to unstable branch localization and stenosis severity estimation. Prior multi-view approaches typically use feature concatenation or static fusion under fixed view sets, which struggles with missing or changing views and lacks explicit modeling of cross-projection anatomical correspondences. We propose a multi-view coronary stenosis report generation framework that builds a controllable synthetic multi-view dataset to mitigate scarce paired samples and fine-grained annotations. It also introduces an Anatomy-Correspondence Fusion module that learns inter-view anatomical correspondences via pose-aware cross-view attention to align and fuse complementary branch evidence, improving branch-level localization consistency and the accuracy of stenosis count and severity prediction. We further design coronary-report-oriented fine-grained metrics covering vessel localization, stenosis count, and severity. Experiments demonstrate state-of-the-art performance on both standard NLG metrics and clinically specific metrics, validating the value of explicit anatomical correspondence modeling for multi-view coronary report generation. The code will be publicly available upon publication.
\end{abstract}

\section{Introduction}
Coronary artery stenosis is a major driver of coronary artery disease and requires precise localization and grading~\cite{Ahmadi2019,Mensah2023}. Projection-based X-ray imaging compresses 3D anatomy into a 2D plane, removing depth cues and introducing view-dependent distortion~\cite{Green2005,preuhs2018viewpoint}. In coronary angiography, thin vessels, uneven contrast, and device interference further exacerbate projection-induced branch overlap and foreshortening, causing lesion visibility to vary substantially across views~\cite{Chen2000,Tu2011}. Near vascular bifurcations, similar local appearance combined with frequent overlap makes single-view branch identification particularly unreliable~\cite{Lee2011,ghobrial2021new,kovcka2020optimal}. Reliable stenosis reporting therefore requires integrating evidence across projections and reasoning over consistent vascular topology. Despite this clinical reality, automated multi-view report generation in coronary angiography remains largely underexplored.

Related work has explored multi-view report generation and fusion mainly in chest X-ray settings, typically fusing images via attention or contrastive objectives and learning cross-view interactions implicitly from the end-task loss. For example, multi-view contrastive learning can strengthen representations and provide auxiliary guidance for transformer decoding~\cite{Bai2025,liu2025enhanced}. Other frameworks first derive prior concepts and then decode reports from fused multi-view knowledge~\cite{10356722}. More generally, multi-view fusion designs such as Duoduo CLIP~\cite{leeduoduo} aggregate features with cross-view attention on top of CLIP-like encoders. Despite their progress, these approaches rarely enforce verifiable anatomical correspondence across projections. As a result, attention mechanisms may align visually similar but anatomically unrelated regions, leading to unstable cross-view aggregation under strong overlap, large viewpoint changes, or near bifurcations.

In this work, we propose a multi-view anatomy-correspondence fusion framework that explicitly learns cross-view correspondences to guide evidence fusion for coronary stenosis reporting. Our main contributions are: (1) We reformulate multi-view coronary stenosis reporting as an alignment-constrained aggregation problem, highlighting the necessity of explicit cross-view anatomical correspondence in projection-based imaging. (2) To overcome the absence of verifiable alignment supervision in real angiograms, we introduce a controllable synthetic angiography generation strategy that provides geometry-derived patch-level cross-view correspondence ground truth together with structured stenosis reports. (3) We design a correspondence-driven multi-view framework in which predicted cross-view correspondence matrices explicitly align auxiliary features within the main-view coordinate space prior to aggregation, enabling anatomically consistent multi-view reasoning. We validate this design through synthetic evaluation and zero-shot transfer to real angiograms.

% (1) We build a synthetic multi-view coronary angiography dataset with precise cross-view correspondence ground truth and view labels, paired with structured stenosis reports. (2) We introduce an anatomy-correspondence module that learns alignment as a supervised and inspectable intermediate representation, rather than an implicit by-product of end-task training. (3) We design correspondence-driven fusion, where learned correspondences directly control cross-view evidence alignment and aggregation before decoding, improving robustness to overlap, foreshortening, and large viewpoint changes. We train on the synthetic dataset and assess transfer to real cases, suggesting generalization beyond simulation.

\section{Method}
\begin{figure*}[t]
\centering
\includegraphics[width=\textwidth]{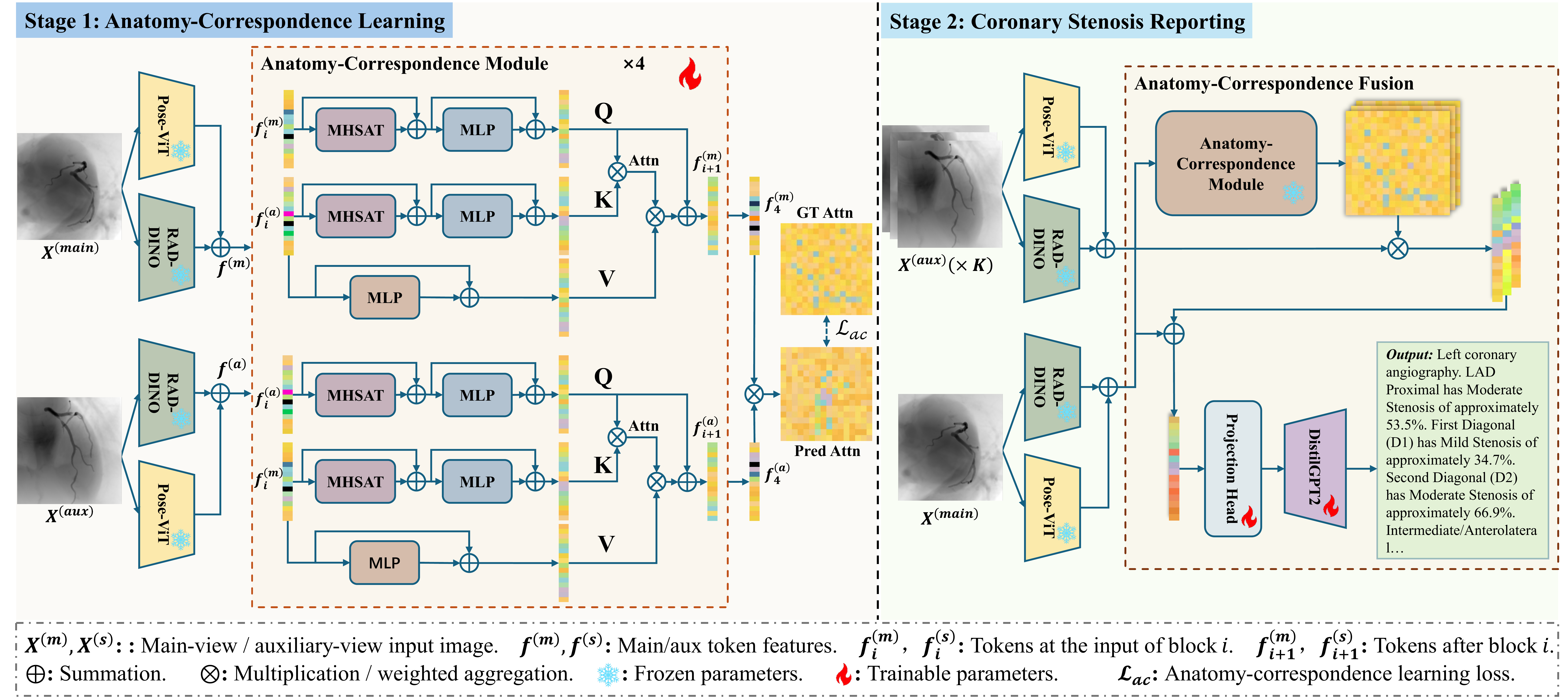}
\caption{Overview of MAC-XA. Stage~1 pretrains ACM with geometry-derived correspondence supervision. Stage~2 uses the learned correspondences for multi-view fusion and structured stenosis report generation.}
\label{fig:overview}
\end{figure*}

An overview of the proposed method is illustrated in Fig.~\ref{fig:overview}, comprising Anatomy-Correspondence Module (ACM) pretraining with geometry-derived correspondence supervision and correspondence-guided multi-view fusion for stenosis report generation.

% \begin{figure}[t]
% \centering
%     \includegraphics[width=1.0\textwidth]{figs/overview.pdf}
% \caption{\textbf{Overview of MAC-XA.} MAC-XA first pretrains an anatomy-correspondence module on synthetic multi-view angiograms with geometry-derived supervision, and then uses the learned correspondences for correspondence-guided fusion and structured stenosis report generation.}
% \label{fig:overview}
% \end{figure}

\subsection{Problem Reformulation}
Single-view stenosis reporting maps one angiography view $X$ to a structured report $\mathbf{y}$. In practice, however, each case contains multiple views acquired in the same exam. We therefore consider a training set $\mathcal{D}_{tr}=\{(\mathbf{X}_i,\mathbf{y}_i)\}_{i=1}^{N}$, where $i$ indexes a case, $\mathbf{X}_i=\{X_i^{(v)}\}_{v=1}^{K_i}$ denotes its $K_i$ available views, and $\mathbf{y}_i$ is the corresponding structured report. The goal is to learn a mapping from $\mathbf{X}_i$ to $\mathbf{y}_i$ for each case.

The central challenge is how to fuse evidence across views reliably. To make multi-view fusion anatomically grounded, we introduce cross-view anatomical correspondence as an explicit alignment prior. Each view is encoded into $P$ patch tokens, and for any view pair we estimate a patch-level correspondence matrix supervised by geometry-derived ground truth in the synthetic data. Given a selected main view and one or more auxiliary views, the predicted correspondences are used to align auxiliary features to the main-view patch coordinate space before fusion, so that report generation is based on anatomically matched evidence rather than implicit feature mixing.

% Single-view stenosis reporting learns a mapping $\hat{\mathbf{y}}=f_{\theta}(X)$ from one angiography view $X$ to a structured report $\mathbf{y}$. In practice, each exam (case) contains multiple views. We train on a dataset $\mathcal{D}_{tr}=\{(\mathbf{X}_i,\mathbf{y}_i)\}_{i=1}^{N}$, where $i$ indexes a case, each case provides $K_i$ views $\mathbf{X}_i=\{X_i^{(v)}\}_{v=1}^{K_i}$ acquired in the same exam ($K_i$ may vary), and $\mathbf{y}_i$ is the corresponding structured report. Our goal is to learn a multi-view model $\hat{\mathbf{y}}_i=f_{\theta}(\mathbf{X}_i)$. The central question is how to fuse evidence across views reliably. To make fusion anatomically grounded, we introduce cross-view anatomical correspondences as an explicit alignment prior: each view is encoded into $P$ patch tokens $T_i^{(v)}\in\mathbb{R}^{P\times d}$, and for any view pair $(\alpha,\beta)$ we learn a patch-level predictor $\mathbf{A}^{pred}_{i,\alpha\rightarrow \beta}=g_{\phi}(X_i^{(\alpha)},X_i^{(\beta)})$, supervised by geometric ground truth $\mathbf{A}^{gt}_{i,\alpha \rightarrow \beta}\in\{0,1\}^{P\times P}$ in our synthetic data. For a selected main view $m$ and auxiliary set $\mathcal{A}_i$, the predicted correspondences are used to align and aggregate features via $\tilde{T}_i^{(a\rightarrow m)}=\mathbf{A}^{pred}_{i,m\rightarrow a}T_i^{(a)}$ and $F_i=\mathrm{Fuse}\!\left(T_i^{(m)},\{\tilde{T}_i^{(a\rightarrow m)}\}_{a\in\mathcal{A}_i}\right)$, and the final report is generated by $\hat{\mathbf{y}}_i=h_{\theta}(F_i)$.

\subsection{Anatomy-Correspondence Ground Truth Preparation}
% 如何以合成数据为基础，创建可靠的解剖对齐监督信号，以及如何生成数据集的。

Fine-grained cross-view anatomical alignment is not directly observable in real coronary angiography. To obtain reliable supervision, we therefore synthesize multi-view X-ray angiograms from real ASOCA 3D CTA volumes~\cite{gharleghi2022automated} and record the view label and projection pose for each rendering. Our pipeline follows a standard CTA-to-angiography simulation process. Starting from coronary artery extraction, we extract vessel centerlines, reconstruct vessel surfaces, and simulate localized stenoses by applying shape- and severity-controlled narrowings at diverse surface locations. 
We render 2D digitally reconstructed angiograms (DRRs) via a physics-based projection model~\cite{zhang2024xa}, producing multi-view projections for both LCA and RCA in each simulated case. To link 3D lesion metadata to clinically meaningful text, we label each 3D centerline branch according to SYNTAX Score segment definitions~\cite{Sianos2005Syntax}, which enables automatic generation of structured plain-text reports during simulation. Each simulated case is paired with a structured stenosis report containing branch identity, severity grade, and percentage stenosis. 

We further construct anatomy-correspondence ground truth (AC-GT) from geometric consistency across known projections. Specifically, we sample 3D points $\{p_n\}$ along the vessel centerlines, project them into each view $v$ using the recorded pose parameters, and map each projected point to a RAD-DINO~\cite{perez2025exploring} patch index $\pi^{(v)}(p_n)\in\{1,\dots,P\}$. For any view pair $(\alpha,\beta)$, we define a binary patch-to-patch ground-truth correspondence matrix $\mathbf{A}^{\mathrm{gt}}_{\alpha \rightarrow \beta}\in\{0,1\}^{P\times P}$ by setting $\mathbf{A}^{\mathrm{gt}}_{\alpha \rightarrow \beta}[i,j]=1$ if there exists a sampled point $p_n$ such that $\pi^{(\alpha)}(p_n)=i$ and $\pi^{(\beta)}(p_n)=j$, and $0$ otherwise. If multiple 3D points fall into the same patch in the main view, the corresponding row can contain multiple positive entries in the auxiliary view, reflecting the inherent many-to-many mapping induced by projection. This matrix provides a geometry-derived supervision target for cross-view anatomical consistency at the patch level.

\subsection{Anatomy-Correspondence Learning}
\textbf{Pose-aware embedding.}
Since correspondence must be inferred across views with substantial pose-induced appearance changes, we encode each view $X^{(v)}$ with two frozen encoders. RAD-DINO provides patch tokens $T^{(v)}\in\mathbb{R}^{P\times d}$ (plus a global \texttt{[CLS]} token) for local appearance cues, while a pose-aware encoder (Pose-ViT, ViT-B/32 pretrained on pose classification) outputs a view embedding $e^{(v)}$ that encodes acquisition pose. We inject $e^{(v)}$ into the \texttt{[CLS]} token to form view-conditioned tokens $f^{(v)}\in\mathbb{R}^{(P+1)\times d}$ for subsequent correspondence learning. For each view pair, the selected main and auxiliary representations are denoted by $f^{(m)}$ and $f^{(a)}$, respectively. These are passed to the ACM, a supervised and inspectable module that stacks four mapping blocks to progressively align the two views.
% \textbf{Pose-aware embedding.}
% For each view $X^{(v)}$, we extract complementary representations with two frozen encoders.
% RAD-DINO provides patch-level tokens $T^{(v)}\in\mathbb{R}^{P\times d}$ (plus a global \texttt{[CLS]} token) for local appearance cues.
% In parallel, a pose-aware encoder (Pose-ViT, ViT-B/32 pretrained on pose classification) outputs a view embedding $e^{(v)}$ that encodes the acquisition pose.
% We inject $e^{(v)}$ into the \texttt{[CLS]} token to obtain a view-conditioned global descriptor, and form the per-view tokens
% $f^{(v)}\in\mathbb{R}^{(P+1)\times d}$ for subsequent cross-view modeling.

% We introduce an Anatomy-Correspondence Module (ACM) to explicitly learn cross-view anatomical correspondences as a supervised and inspectable intermediate representation.
% ACM takes main-view and auxiliary-view tokens $f^{(m)},f^{(a)}\in\mathbb{R}^{(P+1)\times d}$ and stacks four mapping blocks ($\times 4$) to progressively align the two views.

\textbf{Bidirectional correspondence reasoning.}
We model cross-view correspondence by performing correspondence estimation and geometry-aware information transfer in both directions within each mapping block. Specifically, the block consists of two symmetric branches, $m\!\rightarrow\!a$ and $a\!\rightarrow\!m$, which share parameters by swapping the roles of the main and auxiliary views. Within each branch, we first apply lightweight intra-view token refinement (self-attention followed by an MLP) to both views, and then perform a cross-view attention step: main-view tokens produce queries, while auxiliary-view tokens produce keys and values, yielding correspondence logits $Z_{m\rightarrow a}\in\mathbb{R}^{(P+1)\times(P+1)}$ and the row-normalized attention weights
$G_{m\rightarrow a}=\mathrm{softmax}(Z_{m\rightarrow a})$, which update the main-view tokens via a residual cross-view message.

\textbf{Correspondence prediction.}
After the iterative mapping, we compute correspondence logits using dot-product similarity between patch tokens:
\begin{equation}
S_{m\rightarrow a}
=
\frac{f^{(m)}_{1:P}\big(f^{(a)}_{1:P}\big)^\top}{\sqrt{d}}
\in\mathbb{R}^{P\times P},
\qquad
\mathbf{A}^{\mathrm{pred}}_{m\rightarrow a}
=
\sigma\!\left(S_{m\rightarrow a}\right)
\in(0,1)^{P\times P},
\label{eq:acm_pred}
\end{equation}
where each row encodes which auxiliary patches correspond to a given main-view patch, allowing one-to-many matches.

\textbf{Supervision and auxiliary objectives.}
We train ACM with three losses. We supervise $S_{m\rightarrow a}$ with the sparse geometry-derived matrix $A^{\mathrm{gt}}_{m\rightarrow a}$ using a focal BCE with logits to address extreme class imbalance.
In addition, we apply an auxiliary row-wise cross-entropy on the patch-to-patch correspondence logits, i.e.,
$Z^{\mathrm{patch}}_{m\rightarrow a}=Z_{m\rightarrow a}[1\!:\!P,1\!:\!P]$ (excluding the \texttt{[CLS]} token).
The target distribution is obtained by normalizing each positive row of $A^{\mathrm{gt}}$, and rows with no positives are ignored.
We average both directions:

\begin{equation}
\mathcal{L}_{\mathrm{attn}}
=
\frac{1}{2}\Big(
\mathrm{CE}_{\mathrm{row}}(Z^{\mathrm{patch}}_{m\rightarrow a}, A^{\mathrm{gt}}_{m\rightarrow a})
+
\mathrm{CE}_{\mathrm{row}}(Z^{\mathrm{patch}}_{a\rightarrow m}, (A^{\mathrm{gt}}_{m\rightarrow a})^\top)
\Big).
\label{eq:acm_attn_ce}
\end{equation}

Finally, we include a negative-mass penalty that discourages high average predicted probability on non-corresponding entries:
\begin{equation}
\mathcal{L}_{\mathrm{neg}}
=
\frac{1}{|\Omega^-|}
\sum_{(i,j)\in\Omega^-}
\sigma(S_{m\rightarrow a}[i,j])^{\rho},
\qquad
\Omega^-=\{(i,j)\mid A^{\mathrm{gt}}_{m\rightarrow a}[i,j]=0\},
\label{eq:acm_negmass}
\end{equation}
with exponent $\rho$.
The overall objective is
\begin{equation}
\mathcal{L}_{ac}
=
\mathcal{L}_{\mathrm{focal}}
+
\lambda_{\mathrm{attn}}\mathcal{L}_{\mathrm{attn}}
+
\lambda_{\mathrm{neg}}\mathcal{L}_{\mathrm{neg}},
\label{eq:acm_total}
\end{equation}

\subsection{Anatomy-Correspondence Fusion and Reporting}
\textbf{Anatomy-Correspondence Fusion.}
MAC-XA first aligns each auxiliary view to the main-view patch coordinate space and then aggregates the aligned evidence for report generation. For each main--aux pair, we apply ACM to obtain mapped tokens $\bar f$ and convert the similarity logits $S_{m\rightarrow a}$ in Eq.~\eqref{eq:acm_pred} into row-normalized alignment weights $W_{m\rightarrow a}$ via a softmax. We then align auxiliary \emph{raw} patch tokens as $\tilde f^{(a)}_{1:P}=W_{m\rightarrow a}f^{(a)}_{1:P}$ and define the mapping update as $\Delta^{(m)}=\bar f^{(m)}-f^{(m)}$. Using token-wise gates $g_{\mathrm{raw}}$ and $g_{\mathrm{map}}$, we update only the patch tokens:
\begin{equation}
\begin{aligned}
f^{(m)}_{1:P} &\leftarrow f^{(m)}_{1:P}
+ g_{\mathrm{raw},1:P}\odot \tilde f^{(a)}_{1:P}
+ g_{\mathrm{map},1:P}\odot \Delta^{(m)}_{1:P}.
\end{aligned}
\label{eq:fusion_hybrid}
\end{equation}
For a multi-view sample, we randomly select one main view and $K\in\{1,2,3\}$ auxiliary views, and apply Eq.~\eqref{eq:fusion_hybrid} sequentially over auxiliaries to obtain fused tokens $f^{\mathrm{fusion}}$.

% \textbf{Anatomy-Correspondence Fusion.}
% The key design principle of MAC-XA is that auxiliary evidence is not fused directly. Instead, each auxiliary view is first aligned to the main-view patch coordinate space through the predicted correspondence matrix, and only then aggregated for report generation. For each main--aux pair, we apply ACM to obtain mapped tokens $\bar f$ and reuse the patch similarity logits $S_{m\rightarrow a}$ as in Eq.~\eqref{eq:acm_pred}, which we normalize with a temperature-softmax to obtain correspondence weights $A_{m\rightarrow a}=\mathrm{softmax}(S_{m\rightarrow a})$. We align auxiliary \emph{raw} patch tokens by correspondence-weighted aggregation $\tilde f^{(a)}_{1:P}=A_{m\rightarrow a} f^{(a)}_{1:P}$, and additionally inject the explicit mapping update $\Delta^{(m)}=\bar f^{(m)}-f^{(m)}$. We combine both signals with token-wise gates $g_{\mathrm{raw}}, g_{\mathrm{map}}\in(0,1)^{(P+1)\times 1}$ and update only patch tokens:
% \begin{equation}
% f^{(m)}_{1:P}\leftarrow f^{(m)}_{1:P}
% + g_{\mathrm{raw},1:P}\odot \tilde f^{(a)}_{1:P}
% + g_{\mathrm{map},1:P}\odot \Delta^{(m)}_{1:P}.
% \label{eq:fusion_hybrid}
% \end{equation}
% For a multi-view sample, we randomly select one main view and $K\!\in\!\{1,2,3\}$ auxiliary views, and apply Eq.~\eqref{eq:fusion_hybrid} sequentially over auxiliaries to obtain the fused tokens $f^{\mathrm{fusion}}$.

\textbf{Structured Stenosis Reporting Head.}
We attach a lightweight projection head that maps $f^{\mathrm{fusion}}$ to the language-model embedding space and use a cross-attention decoder (DistilGPT2~\cite{radford2019language}) to generate reports under a fixed template and controlled vocabulary. Training uses token-level cross entropy, with optional lexical reweighting following the lexical weighted loss in~\cite{zhang2025semantic}, where we upweight clinically critical tokens specific to coronary stenosis reporting, mainly branch names and stenosis severity levels.

\subsection{Implementation Details}
We synthesize 68,096 DRR angiograms from 38 patients (9,728 simulated cases; 8 LCA and 6 RCA views) and use a patient-wise split of 32/3/3 for training, validation, and testing. We pretrain ACM with AdamW using a learning rate and weight decay of $1\times10^{-4}$. The training objective combines focal BCE ($\alpha=0.75$, $\gamma=2.0$), attention CE ($\lambda_{\mathrm{attn}}=0.05$), and negative-mass regularization ($\lambda_{\mathrm{neg}}=0.05$, $\rho=1.0$), with the latter two terms linearly warmed up. For reporting, we fuse one main view with $K\in\{1,2,3\}$ auxiliary views (random during training and deterministic during evaluation) and optimize only the decoder, projection head, and fusion gates with AdamW at a learning rate of $1\times10^{-4}$.

% \subsection{Implementation Details}
% We use a synthetic dataset of 38 patients (18 diseased, 20 normal). For each patient, we simulate both coronary systems (LCA/RCA) with 128 stenosis configurations per system, varying lesion location, count, and severity, each paired with a structured stenosis report. We render 8 views for LCA and 6 views for RCA, resulting in 9,728 simulated instances and 68,096 DRR angiograms (38,912 LCA, 29,184 RCA). We adopt a patient-wise grouped split (32/3/3 patients for train/val/test) to avoid leakage across simulations and views.

% We follow the two-stage training protocol described above. For downstream fusion, we select one main view and $K$ auxiliary views with $K\in\{1,2,3\}$; view selection is random during training and deterministic for validation/test. ACM pretraining uses AdamW (lr $1\times10^{-4}$, weight decay $1\times10^{-4}$) with focal BCE supervision ($\alpha=0.75,\gamma=2.0$) and two auxiliary terms: attention CE ($\lambda_{\mathrm{attn}}=0.05$) and negative-mass regularization ($\lambda_{\mathrm{neg}}=0.05$, $\rho=1.0$), both linearly ramped up during training. For report generation, we optimize only the decoder, projection, and fusion gates using AdamW with lr $1\times10^{-4}$.

\section{Experiments and Results}
\subsection{Dataset and Evaluation Metrics}
For external verification, we use the public multi-view coronary angiography dataset released by Mahmoudi \textit{et al.}~\cite{Mahmoudi2025}. The release provides angiograms together with a per-case \texttt{SYNTAX-info} dictionary, from which we parse lesion location and stenosis severity and convert them into our structured report template. An experienced clinician applies predefined eligibility criteria to identify 50 evaluable cases, requiring an interpretable \texttt{SYNTAX-info} record, cross-view consistency within the same coronary system, and sufficient vessel visibility across views. Studies with corrupted metadata, mismatched coronary systems across views, severe device occlusion, or major image-quality limitations are excluded.
% For external verification, we use a public multi-view coronary angiography dataset from Mahmoudi \textit{et al.}~\cite{Mahmoudi2025}. The release provides angiograms together with a per-case \texttt{SYNTAX-info} dictionary, we extract lesion location and stenosis severity fields from this dictionary and convert them into our structured stenosis report template. An experienced clinician then performs quality control and selects 50 multi-view cases whose derived structured labels (from \texttt{SYNTAX-info}) match the visible lesions and whose angiograms have sufficient diagnostic quality across views, excluding studies dominated by devices or with poor vessel visibility.

To assess correspondence quality where ground-truth correspondences are available, we measure the agreement between the predicted correspondence confidence map $A^{\mathrm{pred}}\in[0,1]^{P\times P}$ and a sparse binary correspondence matrix $A^{\mathrm{gt}}\in\{0,1\}^{P\times P}$ using Soft-Dice, IoU@Top-$k$, and AUPRC; on real cases without $A^{\mathrm{gt}}$, we rely on clinician review of correspondence correctness. For report generation, we report natural language generation (NLG) metrics (CIDEr, ROUGE-L (R-L)) and template-aware structured metrics by parsing each lesion entry into $I=(b,c,g,p)$ (branch name, count, severity grade, percent stenosis): $\mathrm{F1}_{b}$ requires exact segment-level matching (correct branch \emph{and} SYNTAX-defined sub-segment), $\mathrm{F1}_{s}$ evaluates branch correctness under a coarser grouping that ignores within-branch sub-segments, $\mathrm{F1}_{bn}$ measures stenosis-count consistency on matched branches, and Sev.\ measures severity agreement on BN-matched items.

For reference, Table~\ref{tab:single_view_scores} reports the absolute single-view baseline scores on both synthetic and real data; all subsequent result tables report improvements $\Delta$ over this baseline unless stated otherwise.

\begin{table}[!t]
\caption{Single-view baseline scores (in \%) on synthetic and real datasets.}
\label{tab:single_view_scores}
\centering
\fontsize{8}{9}\selectfont
\setlength{\tabcolsep}{3.6pt}
\renewcommand{\arraystretch}{1.10}
\begin{tabular}{c|cccc|cc}
\hline
\multirow{2}{*}{Setting} &
\multicolumn{4}{c|}{Clinical (Abs.)} &
\multicolumn{2}{c}{NLG (Abs.)} \\
\cline{2-7}
& $\mathrm{F1}_{b} \uparrow$ & $\mathrm{F1}_{\mathrm{bn}} \uparrow$ & $\mathrm{F1}_{s} \uparrow$ & Sev.~$\uparrow$ & CIDEr $\uparrow$ & R-L $\uparrow$ \\
\hline
Single-view (Syn)  & 59.46 & 31.11 & 72.31 & 51.71 & 58.90 & 61.97 \\
Single-view (Real) & 26.16 & 23.51 & 45.18 & 45.80 &  6.49 & 41.39 \\
\hline
\end{tabular}
\end{table}
% \begin{table}[t]
% \caption{Synthetic report generation results: attention--AC-GT alignment (Abs., $\times 100$) and absolute improvements over the single-view baseline ($\Delta$, $\times 100$).
% We report $\Delta=\text{Multi-view}-\text{Single-view}$; positive is better. Best in each column is in \textbf{bold}.}
% \label{tab:metrics_main_delta}
% \centering
% \fontsize{8}{9}\selectfont
% \setlength{\tabcolsep}{3.2pt}
% \renewcommand{\arraystretch}{1.10}

% \begin{tabular}{c|rrr|rrr|rr}
% \hline
% \multirow{2}{*}{Method} &
% \multicolumn{3}{c|}{Alignment (Abs.)} &
% \multicolumn{3}{c|}{Clinical metrics ($\Delta$)} &
% \multicolumn{2}{c}{NLG metrics ($\Delta$)} \\
% \cline{2-9}
% & Dice & IoU@k & AP
% & $\mathrm{F1}_{b}$ & $\mathrm{F1}_{\mathrm{BN}}$ & Sev.
% & CIDEr & R-L \\
% \hline
% MeanPool   & -- & -- & -- & -2.18 & -2.89 & -0.58 & +2.92 & +1.01 \\
% ConcatTok  & -- & -- & -- & -1.45 & +0.91 & -2.84 & -2.65 & -0.33 \\
% DuoDuo     & 0.09 & 1.20 & 4.53 & -4.79 & -3.36 & +0.51 & -11.13 & -3.06 \\
% CrossAttn  & 0.18 & 0.72 & 3.66 & -2.69 & -0.88 & -2.45 & +5.39 & \textbf{+1.71} \\
% Ours       & \textbf{23.08} & \textbf{35.32} & \textbf{53.34}
%            & \textbf{+6.11} & \textbf{+3.60} & \textbf{+2.57}
%            & \textbf{+9.44} & +1.39 \\
% \hline
% \end{tabular}
% \end{table}

% in preamble:
% \usepackage{graphicx}

\begin{table}[!t]
\caption{Synthetic report generation results: attention--AC-GT alignment (Abs., in \%) and improvements ($\Delta$, in \%) over the single-view baseline; see Table~\ref{tab:single_view_scores}.}
\label{tab:metrics_main_delta}
\centering
\fontsize{8}{9}\selectfont
\setlength{\tabcolsep}{3.1pt}
\renewcommand{\arraystretch}{1.10}

\resizebox{\linewidth}{!}{%
\begin{tabular}{c|ccc|cccc|cc}
\hline
\multirow{2}{*}{Method} &
\multicolumn{3}{c|}{Align (Abs.)} &
\multicolumn{4}{c|}{Clinical ($\Delta$)} &
\multicolumn{2}{c}{NLG ($\Delta$)} \\
\cline{2-10}
& sDice $\uparrow$ & IoU@k $\uparrow$ & AUPRC $\uparrow$
& $\mathrm{F1}_{b}$ $\uparrow$ & $\mathrm{F1}_{\mathrm{bn}}$ $\uparrow$ & $\mathrm{F1}_{\mathrm{s}}$ $\uparrow$ & Sev $\uparrow$
& CIDEr $\uparrow$ & R-L $\uparrow$ \\
\hline
MeanPool   & -- & -- & -- & -2.18 & -2.89 & +0.39 & -0.58 & +2.92 & +1.01 \\
ConcatTok  & -- & -- & -- & -1.45 & +0.91 & -2.04 & -2.84 & -2.65 & -0.33 \\
DuoDuo     & 0.09 & 1.20 & 4.53 & -4.79 & -3.36 & -5.11 & +0.51 & -11.13 & -3.06 \\
CrossAttn  & 0.18 & 0.72 & 3.66 & -2.69 & -0.88 & -0.15 & -2.45 & +5.39 & \textbf{+1.71} \\
Ours       & \textbf{23.08} & \textbf{35.32} & \textbf{53.34}
           & \textbf{+6.11} & \textbf{+3.60} & \textbf{+7.03} & \textbf{+2.57}
           & \textbf{+9.44} & +1.39 \\
\hline
\end{tabular}%
}
\end{table}

% \begin{table}[!t]
% \caption{Synthetic report generation results: attention--AC-GT alignment (Abs., in \%) and and improvements ($\Delta$, in \%) over the single-view baseline, see Table~\ref{tab:single_view_scores}.}
% \label{tab:metrics_main_delta}
% \centering
% \fontsize{8}{9}\selectfont
% \setlength{\tabcolsep}{3.1pt}
% \renewcommand{\arraystretch}{1.10}

% \begin{tabular}{c|ccc|cccc|cc}
% \hline
% \multirow{2}{*}{Method} &
% \multicolumn{3}{c|}{Align (Abs.)} &
% \multicolumn{4}{c|}{Clinical ($\Delta$)} &
% \multicolumn{2}{c}{NLG ($\Delta$)} \\
% \cline{2-10}
% & sDice $\uparrow$ & IoU@k $\uparrow$ & AUPRC $\uparrow$
% & $\mathrm{F1}_{b}$ $\uparrow$ & $\mathrm{F1}_{\mathrm{bn}}$ $\uparrow$ & $\mathrm{F1}_{\mathrm{s}}$ $\uparrow$ & Sev $\uparrow$.
% & CIDEr $\uparrow$ & R-L $\uparrow$ \\
% \hline
% MeanPool   & -- & -- & -- & -2.18 & -2.89 & +0.39 & -0.58 & +2.92 & +1.01 \\
% ConcatTok  & -- & -- & -- & -1.45 & +0.91 & -2.04 & -2.84 & -2.65 & -0.33 \\
% DuoDuo     & 0.09 & 1.20 & 4.53 & -4.79 & -3.36 & -5.11 & +0.51 & -11.13 & -3.06 \\
% CrossAttn  & 0.18 & 0.72 & 3.66 & -2.69 & -0.88 & -0.15 & -2.45 & +5.39 & \textbf{+1.71} \\
% Ours       & \textbf{23.08} & \textbf{35.32} & \textbf{53.34}
%            & \textbf{+6.11} & \textbf{+3.60} & \textbf{+7.03} & \textbf{+2.57}
%            & \textbf{+9.44} & +1.39 \\
% \hline
% \end{tabular}
% \end{table}

\subsection{Evaluation on Synthetic Data}

We evaluate correspondence quality and downstream reporting jointly on synthetic data (Table~\ref{tab:metrics_main_delta}). Our comparison includes attention-based baselines, \emph{Cross-Attention} (CrossAttn) and \emph{Duoduo-style Multi-view Attention} (DuoDuo)~\cite{leeduoduo}, as well as simple fusion variants (MeanPool, ConcatTok). For fair comparison, all methods use identical frozen RAD-DINO features and the same pretrained Pose-ViT priors, and differ only in the fusion module. We report the agreement between the predicted correspondence map and AC-GT as absolute scores in the Align (Abs.) columns, and report-generation gains as $\Delta$ over the single-view baseline on clinical and NLG metrics. Our method achieves higher alignment with geometry-derived correspondences and yields the best improvements in structured correctness, indicating that accurate, inspectable correspondences enable more reliable multi-view evidence aggregation.

% \begin{table}[!t]
% \caption{Absolute improvements ($\Delta$) over the single-view baseline on the real dataset.}
% \label{tab:metrics_real_delta}
% \centering
% \fontsize{8}{9}\selectfont
% \setlength{\tabcolsep}{3.4pt}
% \renewcommand{\arraystretch}{1.10}
% \begin{tabular}{c|rrrr|rrrr}
% \hline
% \multirow{2}{*}{Method} &
% \multicolumn{4}{c|}{Clinical metrics ($\Delta$)} &
% \multicolumn{4}{c}{Text metrics ($\Delta$)} \\
% \cline{2-9}
%  & Hard F1 & BN F1 & Soft F1 & Sev. & BLEU-4 & CIDEr & MTR & ROUGE \\
% \hline
% MeanPool  & +4.99 & +1.66 & +1.67 & +10.67 & +2.28 & -1.36 & +3.59 & +2.36 \\
% ConcatTok & -0.69 & +4.93 & -1.43 & +11.60 & +2.38 & -4.86 & +3.40 & +2.68 \\
% CrossAttn & +3.24 & +2.79 & +1.33 & +10.17 & +1.36 & -4.13 & +1.46 & +2.87 \\
% DuoDuo    & +3.97 & +3.62 & +1.51 & +8.40  & +3.98 & -2.13 & \textbf{+4.34} & +4.70 \\
% Ours      & \textbf{+10.19} & \textbf{+11.23} & \textbf{+10.14} & \textbf{+22.57} &
%             \textbf{+4.98} & \textbf{+2.01} & +2.97 & \textbf{+6.20} \\
% \hline
% \end{tabular}
% \end{table}

\begin{table}[t]
\caption{Real report generation results: improvements ($\Delta$, in \%) over the single-view baseline, see Table~\ref{tab:single_view_scores}.}
\label{tab:metrics_real_delta}
\centering
\fontsize{8}{9}\selectfont
\setlength{\tabcolsep}{3.1pt}
\renewcommand{\arraystretch}{1.10}

\begin{tabular}{c|cccc|cc}
\hline
\multirow{2}{*}{Method} &
\multicolumn{4}{c|}{Clinical ($\Delta$)} &
\multicolumn{2}{c}{NLG ($\Delta$)} \\
\cline{2-7}
& $\mathrm{F1}_{b}$ $\uparrow$ & $\mathrm{F1}_{\mathrm{bn}}$ $\uparrow$ & $\mathrm{F1}_{\mathrm{s}}$ $\uparrow$ & Sev. $\uparrow$
& CIDEr $\uparrow$ & R-L $\uparrow$ \\
\hline
MeanPool  & +4.99 & +1.66 & +1.67 & +10.67 & -1.36 & +2.36 \\
ConcatTok & -0.69 & +4.93 & -1.43 & +11.60 & -4.86 & +2.68 \\
DuoDuo    & +3.97 & +3.62 & +1.51 & +8.40  & -2.13 & +4.70 \\
CrossAttn & +3.24 & +2.79 & +1.33 & +10.17 & -4.13 & +2.87 \\
Ours      & \textbf{+10.19} & \textbf{+11.23} & \textbf{+10.14} & \textbf{+22.57}
          & \textbf{+2.01} & \textbf{+6.20} \\
\hline
\end{tabular}
\end{table}

\begin{figure*}[!t]
  \centering
  \begin{subfigure}[t]{0.245\textwidth}
    \centering
    \includegraphics[width=\linewidth]{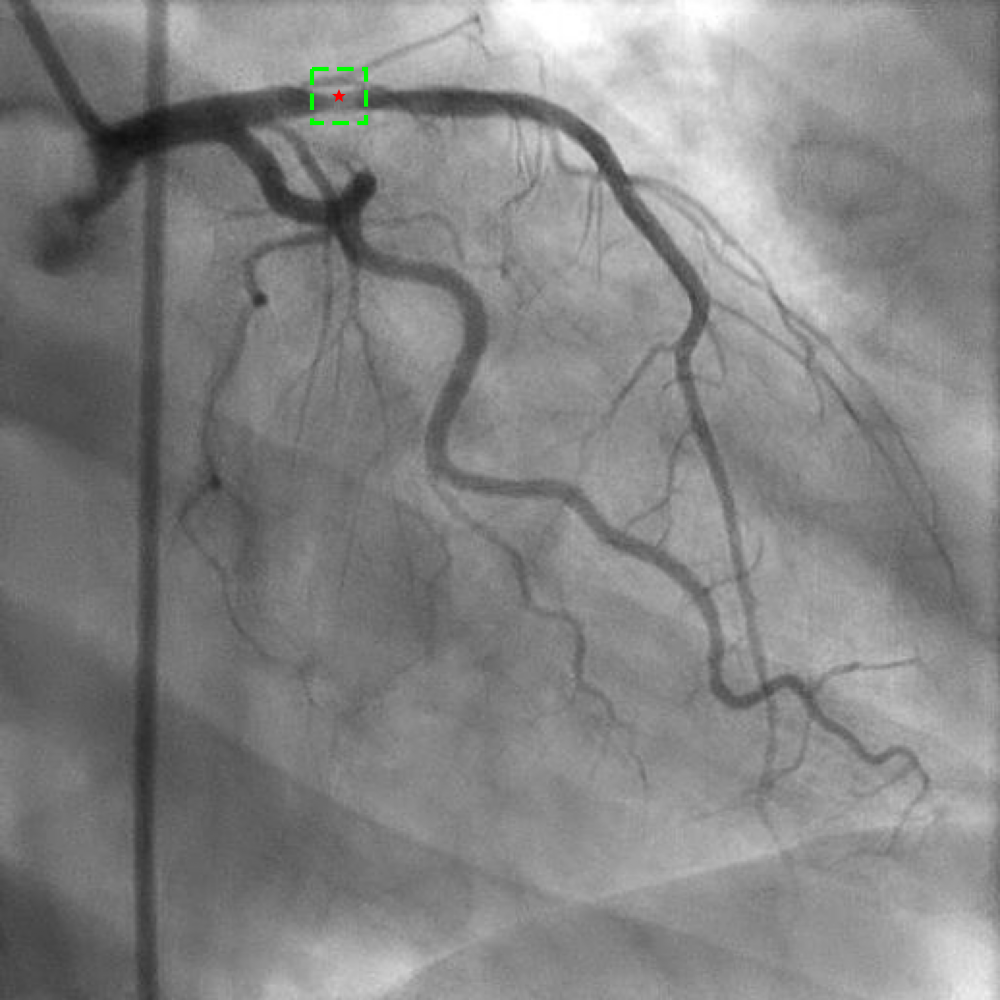}
    \caption{Query}
    \label{fig:real_corr_main}
  \end{subfigure}\hfill
  \begin{subfigure}[t]{0.245\textwidth}
    \centering
    \includegraphics[width=\linewidth]{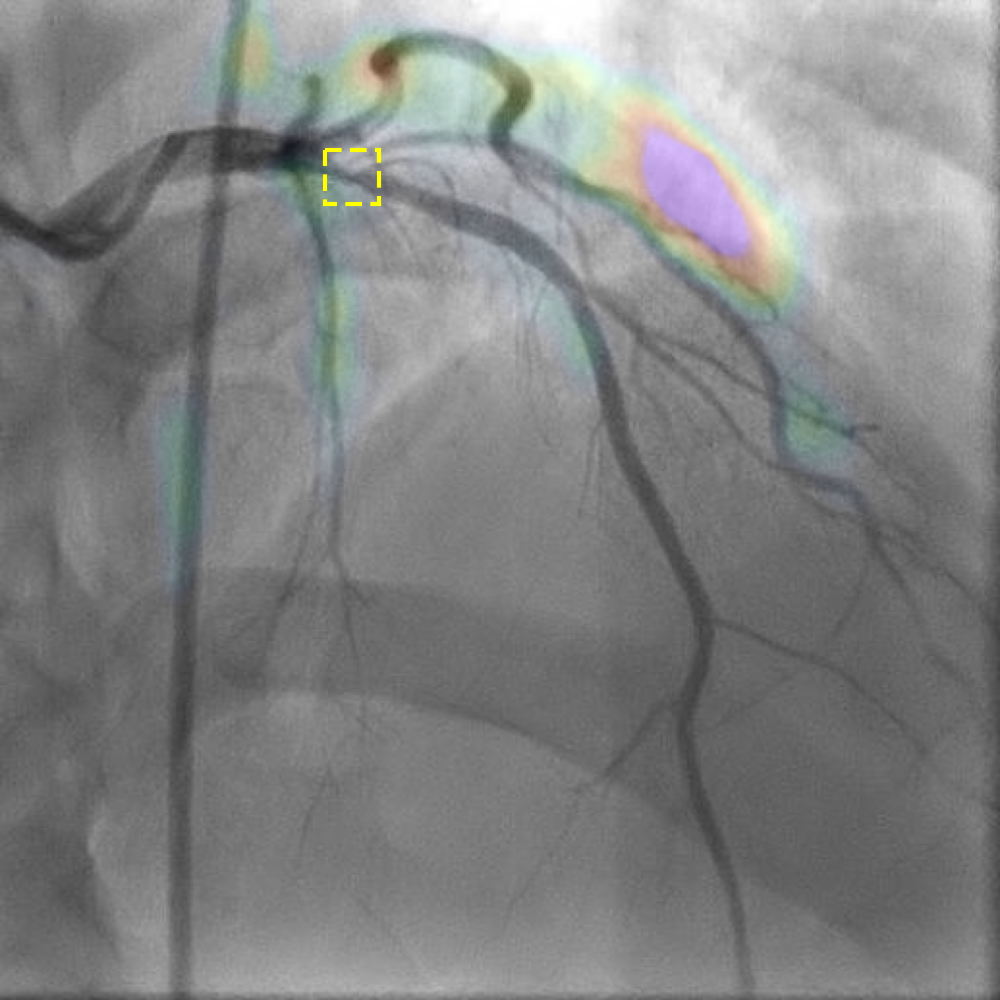}
    \caption{CrossAttn}
    \label{fig:real_corr_cross}
  \end{subfigure}\hfill
  \begin{subfigure}[t]{0.245\textwidth}
    \centering
    \includegraphics[width=\linewidth]{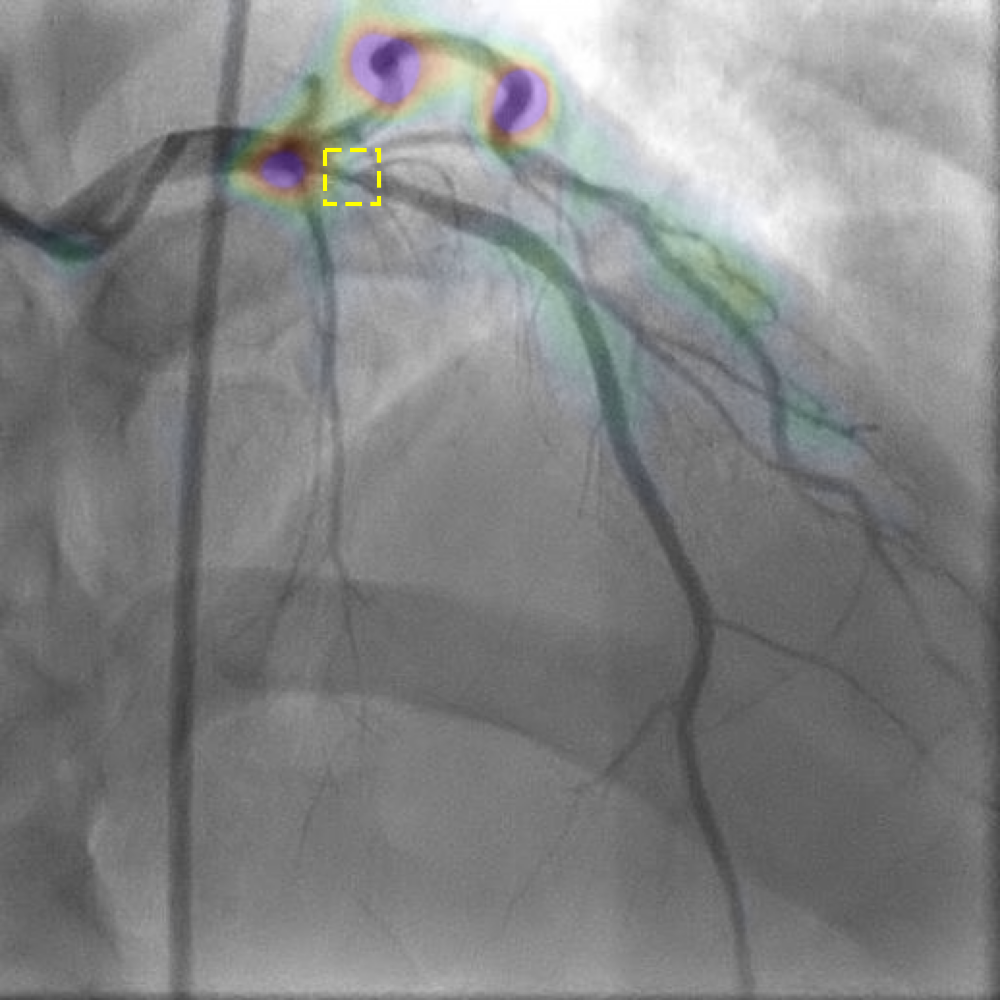}
    \caption{DuoDuo}
    \label{fig:real_corr_duoduo}
  \end{subfigure}\hfill
  \begin{subfigure}[t]{0.245\textwidth}
    \centering
    \includegraphics[width=\linewidth]{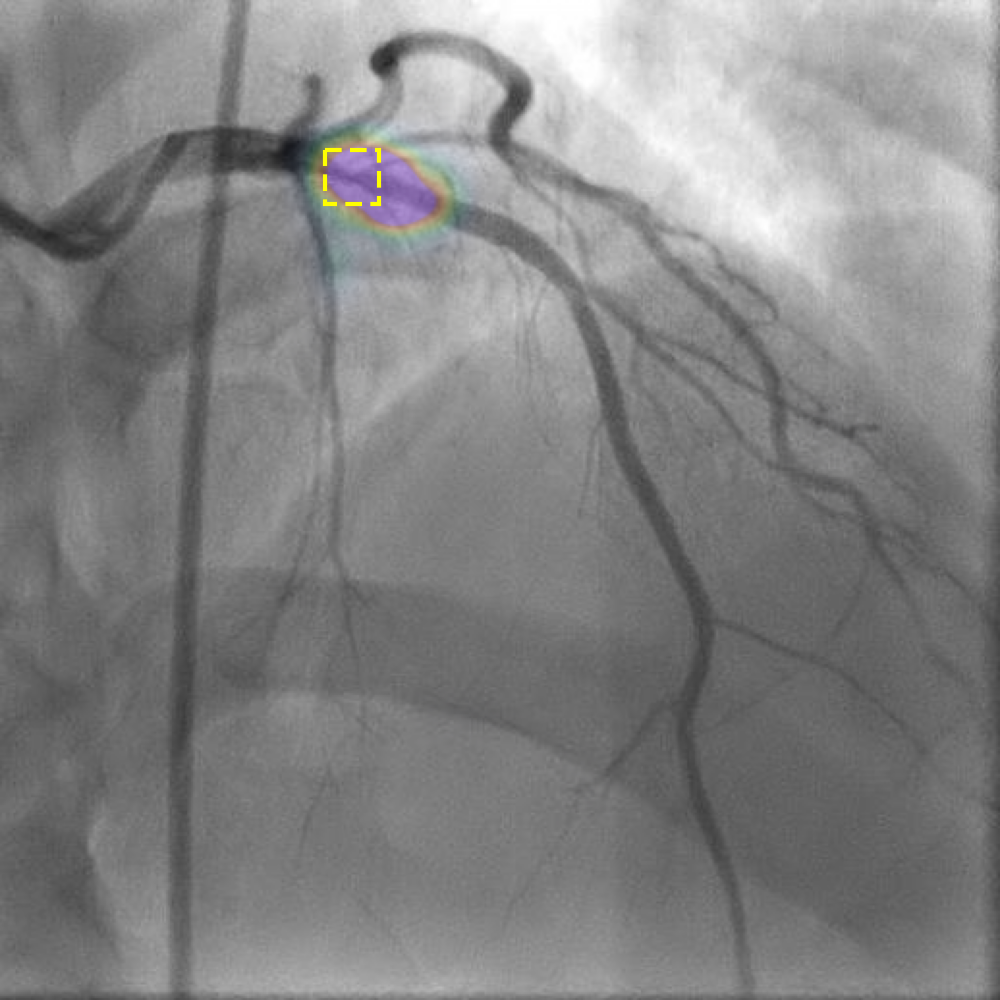}
    \caption{Ours}
    \label{fig:real_corr_ours}
  \end{subfigure}

  \caption{Cross-view correspondence prediction on real angiograms. (a) Reference view with a query patch centered on the stenotic region (green dashed box). (b–d) Predicted correspondence heatmaps on the second view with the manually labeled ground-truth correspondence patch (yellow dashed box).}
  \label{fig:real_corr_vis}
\end{figure*}

\subsection{Zero-shot Real Transfer on 50 Patients}
We evaluate zero-shot transfer by applying models trained on synthetic data to a 50-case real multi-view cohort without further adaptation. Two aspects are assessed. First, we examine bidirectional cross-view correspondence with clinician evaluation. For each case, a stenotic patch in view A is used to predict its correspondence in view B, followed by the reverse direction (A$\rightarrow$B and B$\rightarrow$A), yielding 100 samples. Two experienced clinicians reached consensus that 87/100 predictions were accurately localized. In contrast, CrossAttn and DuoDuo produced diffuse heatmaps with limited anatomical specificity, often highlighting broad vessel regions rather than precise correspondences (Fig.~\ref{fig:real_corr_vis}). Second, we evaluate zero-shot report generation on the same cohort (Table~\ref{tab:metrics_real_delta}). Together, these results demonstrate effective correspondence transfer and improved structured reporting under a clinically realistic, no-adaptation setting.

\section{Discussion and Conclusion}

Interestingly, several multi-view baselines show negative relative gains on synthetic data but positive improvements when transferred zero-shot to real angiograms. We attribute this discrepancy to differences in distribution complexity and information redundancy. The synthetic dataset is generated under controlled conditions with limited noise, minimal occlusion, and stable projection geometry, where single-view inputs already provide sufficient evidence for stenosis assessment. In this regime, naïve multi-view fusion may introduce redundant information or minor misalignment noise, leading to marginal degradation. In contrast, real-world angiography involves incomplete contrast filling, device occlusion, vessel overlap, and projection distortion, making single-view evidence unstable or incomplete. Here, multi-view aggregation becomes beneficial by reducing prediction variance and recovering complementary anatomical cues.

In conclusion, MAC-XA formulates multi-view coronary report generation as an alignment-constrained aggregation problem and shows that geometry-supervised anatomical correspondence learning is key to robust cross-view reasoning. The synthetic dataset plays a critical role by providing controllable, geometry-derived correspondence supervision unavailable in real data, enabling principled pretraining. Together, these findings demonstrate that anatomically grounded alignment, supported by synthetic supervision, is essential for reliable and generalizable multi-view stenosis reporting in clinical practice.

% MAC-XA addresses coronary multi-view report generation as an alignment-constrained aggregation problem, using geometry-supervised correspondence learning to reconcile projection-dependent observations before evidence fusion.

% A key observation is that multi-view fusion helps differently across regimes. On the synthetic set, most multi-view variants perform worse than the single-view baseline, whereas on real angiograms multi-view methods yield consistent gains; notably, MAC-XA improves more on real data than on synthetic. We attribute this to image difficulty: synthetic DRRs are clean and well-opacified, so a single view often suffices, while real angiograms frequently suffer from incomplete contrast injection, catheter/guidewire occlusions, and poor image quality (e.g., low contrast and weak distal-vessel visibility), making complementary views essential for complete evidence capture. Zero-shot transfer to a 50-patient real cohort, supported by clinician scoring, suggests that the learned correspondences generalize beyond simulation.

\bibliographystyle{splncs04}
\bibliography{reference}

\end{document}